\documentclass[lettersize,journal]{IEEEtran}
\usepackage{amsmath,amsfonts}
\usepackage{algorithmic}
\usepackage{algorithm}
\usepackage{array}
\usepackage[caption=false,font=normalsize,labelfont=sf,textfont=sf]{subfig}
\usepackage{textcomp}
\usepackage{stfloats}
\usepackage{url}
\usepackage{verbatim}
\usepackage{graphicx}
\usepackage{cite}

\usepackage{amsmath,amssymb,amsfonts}
\usepackage{xcolor}

\def\BibTeX{{\rm B\kern-.05em{\sc i\kern-.025em b}\kern-.08em
    T\kern-.1667em\lower.7ex\hbox{E}\kern-.125emX}}
\usepackage{comment}
\usepackage{dirtytalk}
\usepackage{multirow}

\begin{document}

\title{A CNN-based End-to-End Learning for RIS-assisted Communication Systems}

\author{{Nipuni  Ginige, Nandana Rajatheva, and Matti Latva-aho}~\\ \IEEEmembership{Center for Wireless Communications,
University of Oulu,
Finland \\
\{nipuni.ginige,  nandana.rajatheva, matti.latva-aho\}@oulu.fi}}



\maketitle

\begin{abstract}
Reconfigurable intelligent surface (RIS) is an emerging technology that is used to improve the system performance in beyond 5G systems. In this letter, we propose a novel convolutional neural network (CNN)-based autoencoder to jointly optimize the transmitter, the receiver, and the RIS of a RIS-assisted communication system. The proposed system jointly optimizes the sub-tasks of the transmitter, the receiver, and the RIS such as encoding/decoding, channel estimation, phase optimization, and modulation/demodulation. Numerically we have shown that the bit error rate (BER) performance of the CNN-based autoencoder system is better than the theoretical BER performance of the RIS-assisted communication systems.  
\end{abstract}


\section{Introduction}
Reconfigurable intelligent surface (RIS) is an emerging technology that is going to be used in beyond fifth-generation (5G) networks. RIS is a planar surface that can be used to improve the performance of the system by changing the phase of the incoming signal especially when the transmitter to receiver direct path is blocked.

Recently deep learning techniques have caught the interest of researchers to apply in wireless communication. Deep learning is used to improve the accuracy of many areas in wireless communication systems such as signal detection \cite{8227772}, channel estimation \cite{8949757} and channel decoding\cite{7926071}. The authors in \cite{9345504} proposed a deep learning receiver (DeepRx), which claims a better radio performance than traditional receivers. 

Furthermore, deep learning can be used to jointly optimize the transmitter and the receiver. The authors in \cite{8445920} proposed deep neural network (DNN)-based autoencoders for end-to-end learning of OFDM-based communication systems. A novel convolutional neural network (CNN)-based autoencoder learning system for an intelligent communication system is proposed by the authors in \cite{8755977}. It is beneficial to jointly optimize the transmitter and receiver using deep learning-based autoencoders rather than optimizing separately sub tasks of the transmitter (encoding, modulation) and sub tasks of the receiver (channel estimating, signal detection, channel decoding, demodulation). \cite{8445920,8755977}.

Due to its learning capability and reduced processing complexity, deep learning is used for RIS-assisted communication systems as well. In our previous work, we proposed a deep learning-based channel estimation method for an OFDM-based RIS-assisted system \cite{9569694}. The authors in \cite{khan2019deep} presented a deep learning approach for estimating and detecting symbols in signals transmitted through RIS.  An unsupervised learning-based approach  is proposed by the authors in \cite{8955968}  to find the optimal phase shift of the RIS.

It is shown that joint optimization of the transmitter and receiver in a conventional communication system using deep learning-based autoencoders improves its bit error rate (BER) performance \cite{8445920,8755977}. Furthermore, such a system is much simpler and straightforward than the complex algorithms involved in each sub-task of the signal processing chain. This motivated us to investigate the potential of using autoencoders for end-to-end learning of a RIS-assisted communications system instead of optimizing its sub-tasks in the transmitter, receiver and the RIS such as encoding/decoding, channel estimation, phase optimization of the RIS, and modulation/demodulation separately. 

\subsection{Contributions}
In this letter, we exploit the learning capability of deep learning-based autoencoders for joint optimization of the transmitter, RIS, and the receiver in a RIS-assisted communications system to improve its performance. The main contributions of this letter is as follows:
\begin{itemize}
    \item Novel CNN-based autoencoder (CNN-AE) system for the end-to-end learning of  RIS-assisted systems.
    \item The proposed system can work with perfect channel state information (CSI) and imperfect CSI scenarios.
    \item The proposed end-to-end learning system architecture jointly optimize the tasks of encoding/decoding, channel estimation, phase optimization of the RIS, and modulation/demodulation.
    \item Proposed system can find the optimal reflection coefficients to be used in the data transmission phase with the knowledge of the channel.
\end{itemize}

The rest of the letter is organized as follows. The system model is presented in Section \ref{SM}. The proposed CNN-AE end-to-end learning system is presented in Section \ref{CNN-AE}. The numerical results are presented in Section \ref{results} and Section \ref{conclusion} concludes our letter.

\section{System Model}
\label{SM}
\begin{figure}[ht]  
	\centering
	\includegraphics[width=0.5\textwidth]{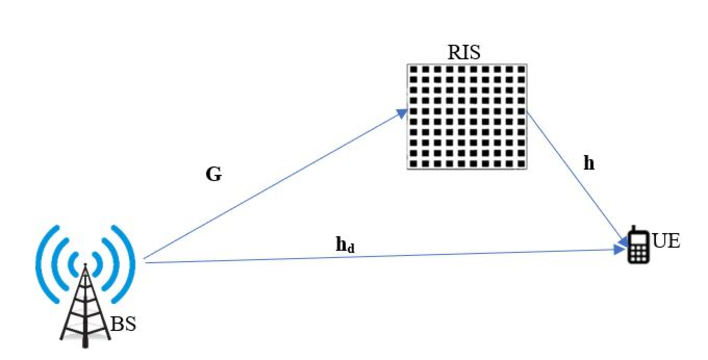}
	\caption{Illustration of the system model.}
	\label{fig:Illustration of the system model}
\end{figure}

We consider a downlink of a RIS-assisted system as shown in Fig. \ref{fig:Illustration of the system model}. A RIS is placed in between the base station (BS)  and the user equipment (UE).  We assume single antenna is used at the BS and the UE, while the RIS is equipped with $N$ reflecting elements. 

At the transmitter side, the BS chooses a symbol $s$, with $k$ information bits, to transmit to the UE through the cascade channel along BS-RIS-UE and the BS-UE direct channel. BS applies a transformation to the symbol $s$  to generate the transmitted signal $\mathbf{x} \in \mathbb{C}^n$ where $n$ is the number of time slots to transmit the signal. 

Received signals from the BS to RIS is reflected to the UE by the RIS using the calculated optimal phase shifts. Received signal, $\mathbf{y} \in \mathbb{C}^n$, at the UE can be expressed as, 

\begin{equation} \label{eq1}{\mathbf{y}}=  \left( { \mathbf{G} \mathbf{\Theta}\mathbf{h}  + {\mathbf{h}}_{d}} \right) \mathbf{x}+ {\mathbf{w}},\end{equation}
where $\mathbf{G}$ is the channel between BS and RIS, $\mathbf{h}$ is the channel between RIS and UE, $\mathbf{h_d}$ is the direct channel between BS and UE. Moreover, $\mathbf{\Theta} = \text {diag}(\boldsymbol{\theta}) $ is the phase shift matrix of the RIS, where $\boldsymbol{\theta} = [\theta_1, \theta_2, \cdots, \theta_N]^T \in \mathbb{C}^{N \times 1}$ and $|\theta_n|=1$. ${\mathbf{w}} \sim {\mathcal N_{c} }(0 , \sigma ^{2})\in \mathbb{C}^n$ denotes the additive white Gaussian noise (AWGN) with noise power of $\sigma^2$. Therefore, the received signal-to-noise ratio (SNR) at the UE is as follows:
\begin{equation} \gamma = \frac {1}{\sigma ^{2}}|(\mathbf{G}\mathbf{\Theta}\mathbf{h} +\mathbf{h_d})^{T}|^{2}. \end{equation}

Optimal phase shifts $\boldsymbol{\theta}$ can be found by maximizing the SNR at UE, by solving the following optimization problem.  
\begin{align} \label{opt}  \qquad&\underset {\boldsymbol {\theta }}{\text {max}}\qquad \left \|{\mathbf{G} \mathbf{\Theta}\mathbf{h}+\mathbf{h_d}}\right \|^{2} \notag \\&~\, \text {s.t.}\qquad \, |\theta _{n}|=1, \; n = 1, \ldots, N.\end{align}

Numerical solutions for the problem \eqref{opt} only provides sub-optimal solutions \cite{8955968}. Therefore, we solve the problem \eqref{opt} using a deep learning  based approach.

\begin{figure}[ht]  
	\centering
	\includegraphics[width=0.5\textwidth]{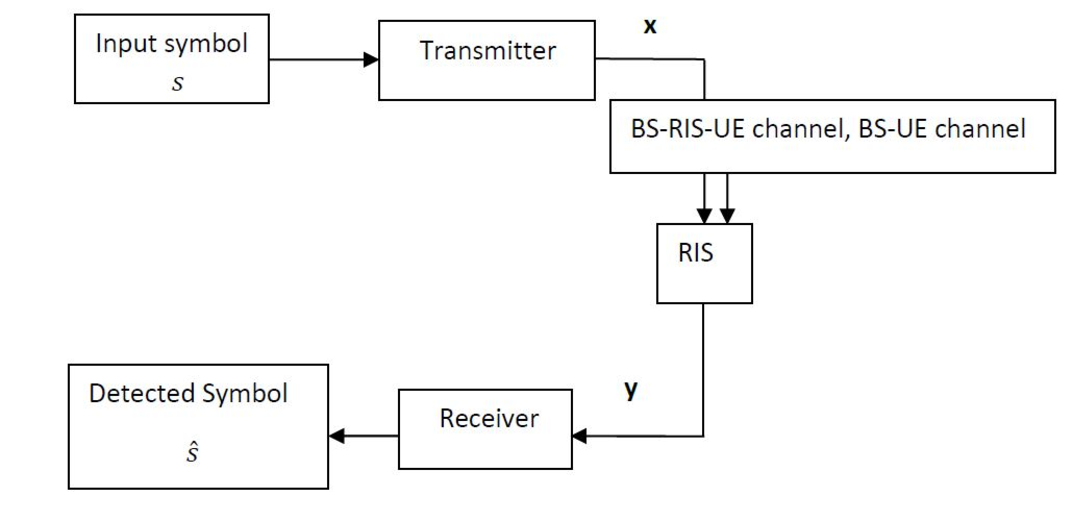}
	\caption{Illustration of the system architecture.}
	\label{fig:Illustration of the system architecture}
\end{figure}

At the UE, using $\mathbf{y}$, receiver decodes the estimate $\hat{s}$ of the original symbol $s$. The above mentioned system architecture is shown in the Fig. \ref{fig:Illustration of the system architecture}.

We use a categorical cross-entropy loss function, ${L}(\phi)
$  with Adam optimizer for optimization of the DNN
parameters.

\begin{equation}
   {L}(\phi)=-\frac {1}{M}\sum _{i=1}^{M} s_i \log{\hat{s}_i},
\end{equation}
where $M$ is the batch size.

\section{CNN-AE learning system}
\label{CNN-AE}
\subsection{CNN-AE learning system with perfect CSI}

\label{withcsi}
\begin{figure*}[ht]  
	\centering
	\includegraphics[width=1\textwidth]{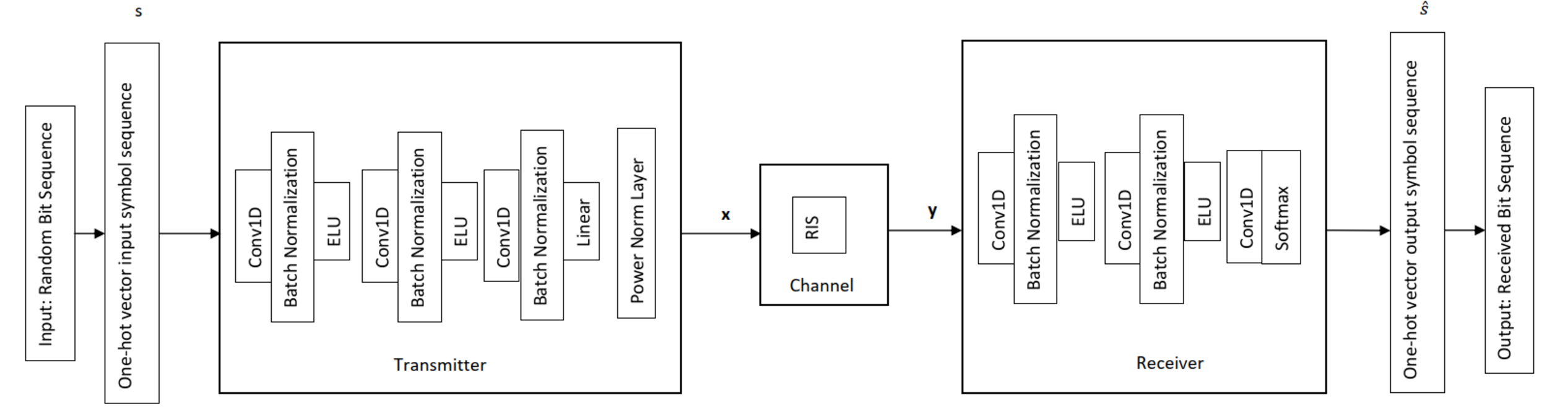}
	\caption{CNN - based end-to-end learning system.}
	\label{fig:CNN}
\end{figure*}

\begin{figure}[ht]  
	\centering
	\includegraphics[width=0.5\textwidth]{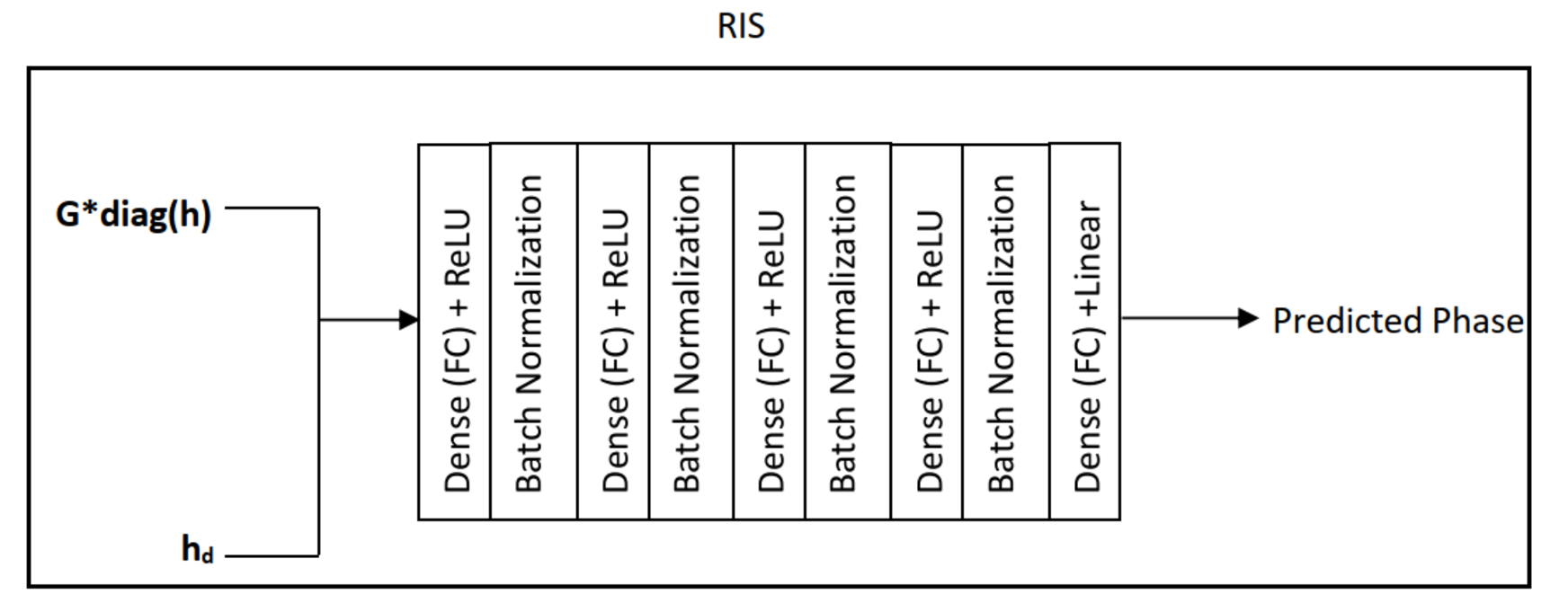}
	\caption{RIS block.}
	\label{fig:RIS}
\end{figure}

 The proposed CNN-based end-to-end learning for RIS-assisted wireless system is shown in Fig. \ref{fig:CNN} and the RIS block of it is shown in Fig. \ref{fig:RIS}.
 
 The detailed description of Fig. \ref{fig:CNN} is as follows: The transmitter block consists of two sets of 1-dimensional convolutional (Conv1D) layer, batch normalization (BN) layer, and exponential linear unit (ELU) activation layer as shown in Fig. \ref{fig:CNN}. Then the last set consists of Conv1D layer, BN layer, and linear activation layer. Those three sets are followed by a power normalization layer to satisfy the power constraint of the transmitted signal. The transmitter is responsible for the encoding and modulation of the input symbol sequence.
 
The channel block consists of the RIS block and the noise layer. Generally,  the channel layer can be described as the conditional probability density function $p(\mathbf{y} |\mathbf{x} )$. The transmitted signal $\mathbf{x} $ is going through the RIS channel which consists of the direct channel and the cascade channel. Both the direct channel and the cascade channel are Rayleigh channels. When the knowledge of the direct channel and cascade channel are given to the RIS block, it calculates the predicted optimal phase shift which maximizes the SNR according to \eqref{opt}.

 The RIS block is pre-trained for random RIS channels and the pre-trained model is used to find the optimal phase shift during the training process and the testing process of the CNN-AE. As shown in Fig. \ref{fig:RIS}, the RIS block consists of four sets of fully connected (FC) dense layers with rectified linear unit (ReLU) activation layers and BN layers. Finally, those four sets are followed by a FC dense layer with a linear activation layer. The effective channel of the RIS which consists of the cascade channel and the direct channel is convolved with the transmitted signal $\mathbf{x} $ and an AWGN with fixed variance $\sigma = (2RE_b/N_0)^{-1}$ is added to the signal. Here, $R=k/n$(bits/channel use) is the code rate and $E_b/N_0$ is the SNR.

Input for the receiver block is the received signal $\mathbf{y} $ and the CSI of the direct and cascade channels. The receiver consists of two sets of Conv1D layer, BN layer, and ELU activation layer as shown in Fig. \ref{fig:CNN}. And finally a Conv1D layer and a softmax activation layer. The responsibilities of the receiver are the decoding and the demodulation. It decodes the received signal $\mathbf{y} $ out of $2^k$ possibilities using the learned system.

The layout of the proposed CNN-AE end-to-end learning for the RIS-assisted communication system is presented in Table \ref{table:1}.

\begin{table}[ht]
\centering
\caption{Layout of the proposed CNN- based AE.}
\label{table:1}
\begin{tabular}{| c |c| c |}
    \hline
    Block&  & Output Dimension \\ 
    \hline
    \multirow{3}{7em}{Transmitter} & One-hot input & $ 2^k$ \\ 
    & Conv1D + BN + ELU &  $256$\\
    & Conv1D + BN + ELU & $ 256$\\
    & Conv1D + BN + Linear& $ 2n$\\
    & Power Norm Layer & $ 2n$\\
    \hline
    \multirow{5}{7em}{RIS + AWGN } &Dense (FC) + ReLU + BN&$32 \times N \times  2n$ \\
    & Dense (FC) + ReLU + BN&$16 \times N\times 2n$\\
    & Dense (FC) + ReLU + BN&$8 \times N\times 2n$\\
    & Dense (FC) + ReLU + BN&$4 \times N\times 2n$\\
    & Dense (FC) + Linear&$N\times  2n$\\
    & AWGN & $ L \times 2n$\\
    \hline
    \multirow{3}{7em}{Reciever } &Conv1D + BN + ELU & $256$\\
    &Conv1D + BN + ELU &$256$\\
    &Conv1D + BN + Softmax &$ 2^k$\\
    \hline
\end{tabular}

\end{table}

\subsection{CNN-AE learning system with channel estimation}

 \begin{figure}[ht]  
	\centering
	\includegraphics[width=0.5\textwidth]{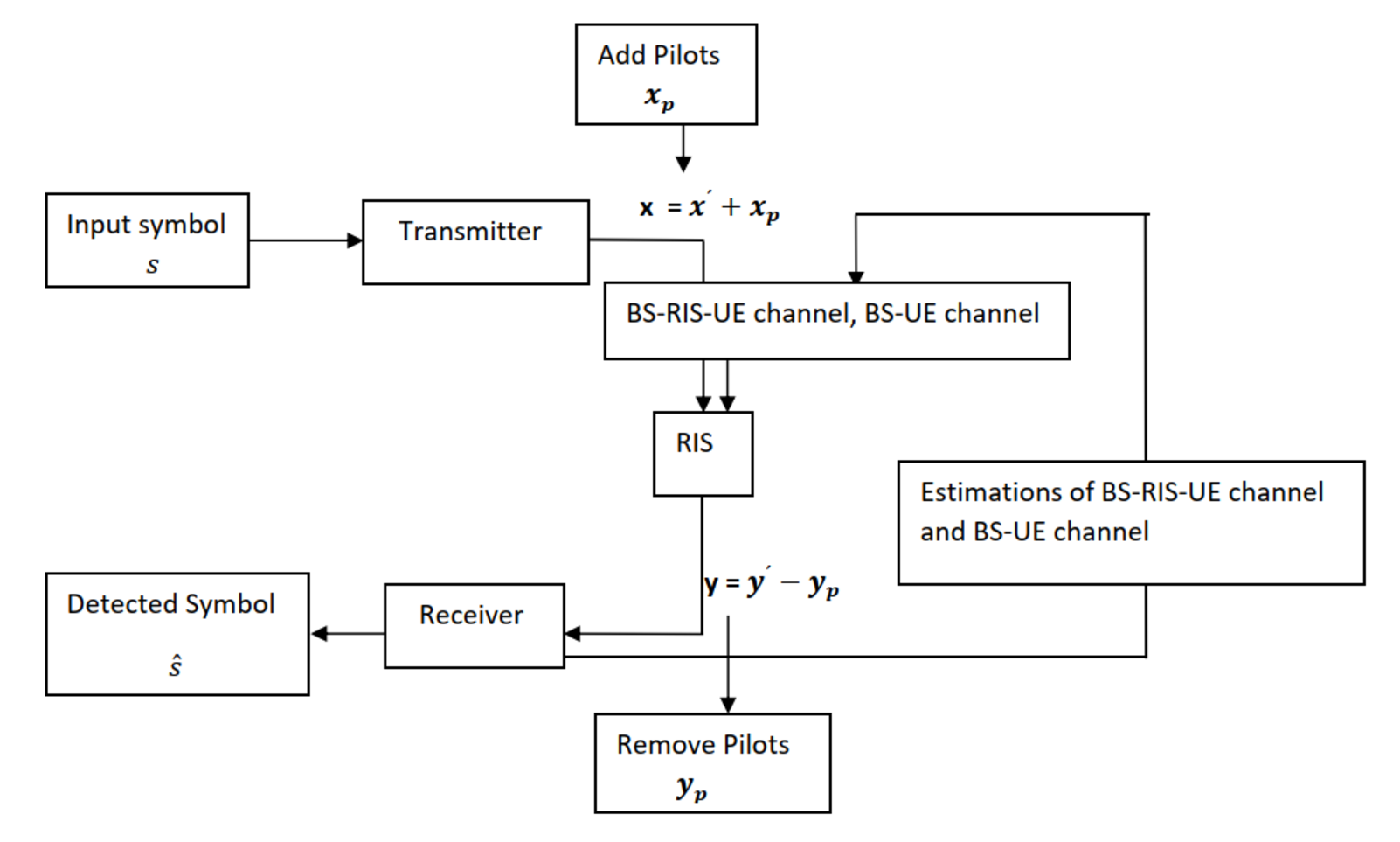}
	\caption{Illustration of the system architecture with channel estimation.}
	\label{fig:Illustration of the system architecture2}
\end{figure}

The system architecture of the autoencoder learning system with channel estimation is shown in Fig. \ref{fig:Illustration of the system architecture2}. The pilot symbols are added to the output of the transmitter. We consider  $T$ consecutive pilot symbols which are equal to $N+1$ (number of elements in the RIS +1). During the transmission of pilot symbols, we consider a predefined DFT-based reflection coefficient matrix for the RIS. Estimated channels are feedbacked to the RIS and they are used to find the optimal reflection coefficients for the data transmission.

The received signal related to the pilot symbols can be written as follows:
\begin{equation} \label{eq4}{\mathbf{y}_p}=  \left( { \mathbf{G} \mathbf{\Theta}\mathbf{h}  + {\mathbf{h}}_{d}} \right) \mathbf{x}_p+ {\mathbf{w}_p},\end{equation}
Hence, the least-square estimation of the effective channel can be written as
\begin{align} \label{eq5} {\mathbf {\hat{H}}^{(LS)}}=\left({\mathbf {x}}_p \right )^{-1}{\mathbf {y}}_p = \mathbf{H}_p + \left({\mathbf {x}}_p \right )^{-1}\mathbf{w}_p.\end{align}
where $ \mathbf{H}_p = { \mathbf{G} \mathbf{\Theta}\mathbf{h}  + {\mathbf{h}}_{d}} $. Since a predefined reflection coefficient matrix is used during the pilot transmission, we can find the estimations for the channels. After getting the knowledge of the channels using channel estimation, rest of the procedure to decoded the symbols are same as the section \ref{withcsi}.


\section{Numerical Results }
\label{results}
In this section, we evaluate the performance of the CNN-AE system for the end-to-end learning of the RIS-assisted communication systems. 

We generate random binary bit sequences consisting of 0's and 1's which are drawn from uniform distribution as training, validation, and testing data sets. The CNN-AE system is trained with 1280000 data symbols where each symbol has $k$ bits. From that training data set 20\% of data symbols are used for validation. The system is tested with 3200000 data symbols. Batch size for training is 128 and batch size  for testing is 64.

We define the loss function as the categorical cross-entropy loss function between one-hot vectors of the input symbol sequence and the output symbol sequence. We use the Adam optimizer with an initial learning rate of 0.001. Also, we set the early stopping with patience 100. When the validation loss of a consecutive 50 epoch remains the same, the learning rate is reduced by a 0.1 factor. The system is trained using 150 epochs. The model is trained at a fixed $E_b/N_0$ value of 16dB. The proposed end-to-end learning system can be trained at a specific SNR while testing across the whole range. 

During the pre-training phase of the RIS block, we use the Adam optimizer with an initial learning rate of 0.001. Also, we set the number of maximal epochs to 1000 and early stopping with patience 20. When the validation loss of a consecutive 10 epoch remains the same, the learning rate is reduced by a 0.33 factor.

\subsection{CNN-AE learning system with perfect CSI}

\begin{figure}[ht]  
	\centering
	\includegraphics[width=0.5\textwidth]{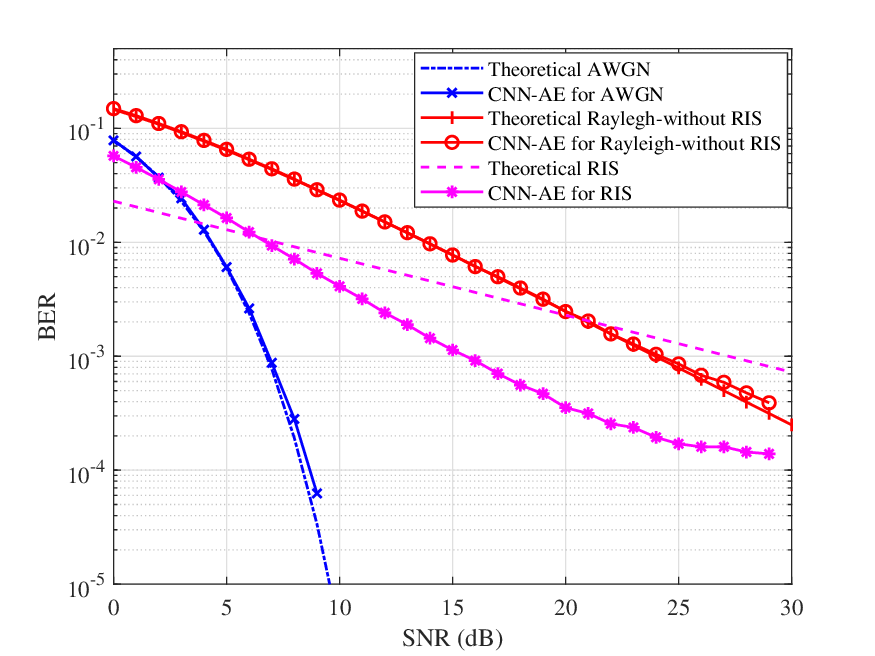}
	\caption{BER performance of the CNN-AE system (BPSK).}
	\label{fig:SNR vs BER}
\end{figure}

In this section we assume perfect CSI. Fig. \ref{fig:SNR vs BER} shows the BER performance of the CNN-AE system for various SNR values. The authors in \cite{8755977}, have numerically shown that the achieved BER performance of the CNN-AE system is matched with the theoretical BPSK BER performance for AWGN and Rayleigh channels.

Theoretically,  RIS-assisted systems with Rayleigh channels show better BER performance compared to the Rayleigh channels without RIS. According to Fig. \ref{fig:SNR vs BER}, CNN-AE system for RIS-assited system shows better BER performance compared to the the theoretical BER performance of RIS-assisted system with BPSK modulation scheme. We assume $N=16$ number of elements in the RIS for this simulation. The theoretical BER values of the RIS-assisted system is calculated using equations in \cite{8796365}.

\begin{figure}[ht]  
	\centering
	\includegraphics[width=0.5\textwidth]{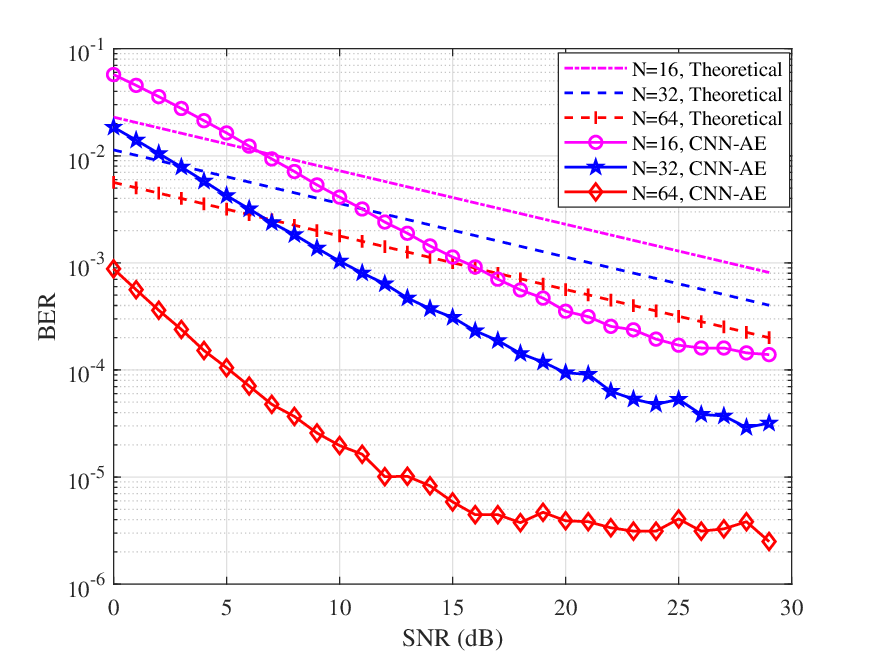}
	\caption{BER performance (BPSK) of proposed RIS assisted CNN-based AE system vs theoretical \cite{8796365} for different RIS sizes.}
	\label{fig:comparison}
\end{figure}

The comparison of the BER performance of the CNN-AE for the RIS assisted system for a different number of RIS elements is shown in Fig. \ref{fig:comparison}. Here we use BPSK modulation scheme where $k=1$. It also shows the comparison between the CNN-AE system and the theoretical BER performance of the RIS-assisted system with a different number of RIS elements.

Fig. \ref{fig:comparison} shows that the CNN-AE system has better  BER performance for the RIS-assisted system compared to the theoretical BER performance when the number of elements in the RIS are increased. Further, CNN-AE learning system shows significantly low BER values than theoretical values especially at high number of RIS elements. 

\begin{figure}[ht]  
	\centering
	\includegraphics[width=0.5\textwidth]{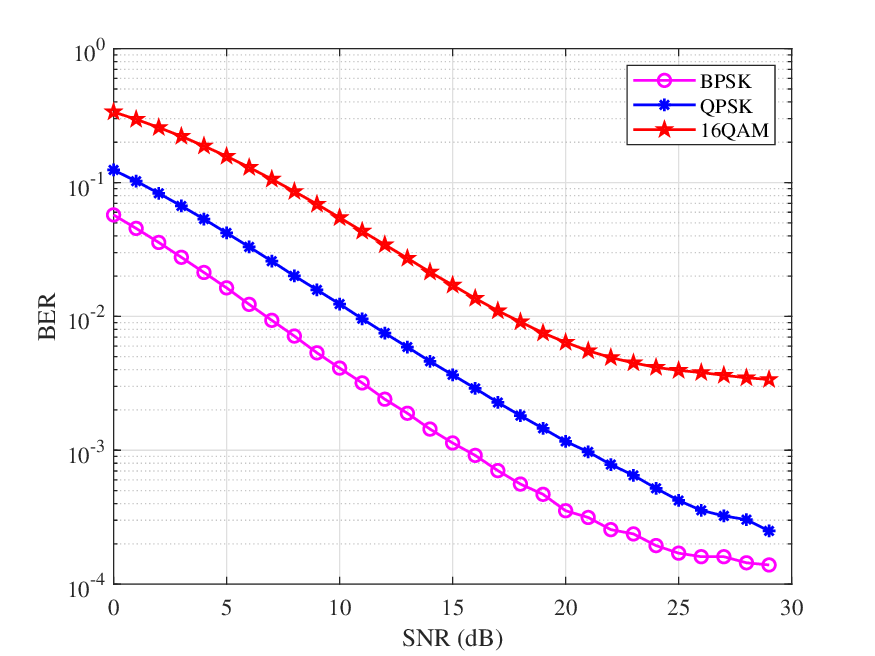}
	\caption{BER performance of the CNN-AE for RIS-assisted system  when compared with corresponding BPSK, QPSK, 16QAM modulations.}
	\label{fig:modcomparison}
\end{figure}

BER performance of the CNN-AE for RIS-assisted system  when compared with corresponding BPSK, quadrature phase shift keying (QPSK), 16 quadrature amplitude modulation (16QAM)  modulations when $N=16$ shows in Fig. \ref{fig:modcomparison}. It proves the compatibility of  our system with different modulation schemes.

\begin{figure}[ht]  
	\centering
	\includegraphics[width=0.5\textwidth]{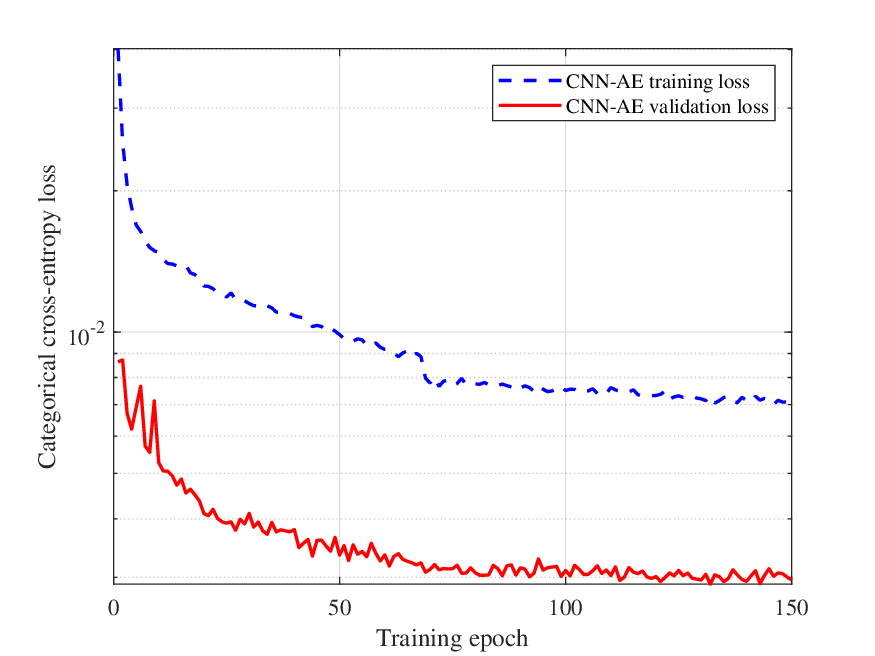}
	\caption{Training loss  and validation loss of the CNN-AE for the RIS-assisted system for BPSK}
	\label{fig:convergence}
\end{figure}

Fig. \ref{fig:convergence} shows the training loss and the validation loss of the CNN-AE end-to-end learning system for the RIS-assisted communication system.  It shows that the CNN-AE learning system has fast convergence since both the training loss and validation loss of the system converge quickly with a small number of epochs.

\subsection{CNN-AE learning system with channel estimation}

\begin{figure}[ht]  
	\centering
	\includegraphics[width=0.5\textwidth]{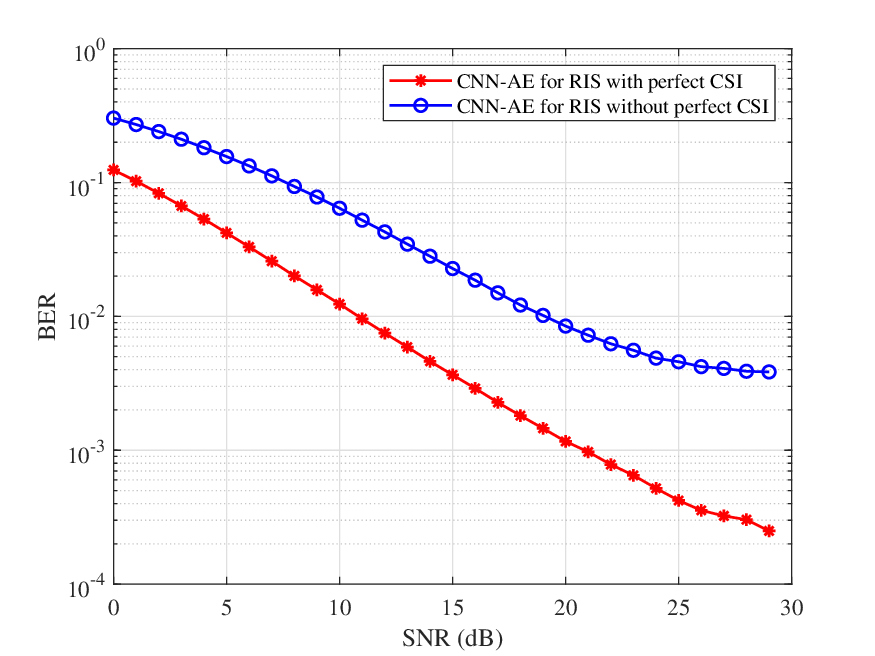}
	\caption{BER performance of the CNN-AE system with and without perfect CSI.}
	\label{fig:SNR vs BER CE}
\end{figure}

Fig. \ref{fig:SNR vs BER CE} compares the BER performance of the CNN-AE learning system for both scenarios with perfect CSI and imperfect CSI. For this simulation we assume QPSK modulation scheme and $N=16$. According to Fig. \ref{fig:SNR vs BER CE}, CNN-AE for RIS-assisted system shows high BER values when there is no perfect CSI knowledge compared to when it has perfect knowledge of CSI. In this simulation we assume least square channel estimation for without perfect CSI scenario.  As our future work, we will further investigate to improve the BER performance for without perfect CSI scenario. Moreover, high accuracy deep learning-based channel estimation algorithms is planned to use instead of least square channel estimation.

\section{Conclusion}
\label{conclusion}

This letter proposed a novel CNN-AE end-to-end learning for RIS-assisted communication systems. RIS is an emerging technology in  beyond 5G systems that is used to improve the performance of communication systems. We propose a CNN-AE end-to-end learning system to jointly optimize the transmitter, RIS, and the receiver of the RIS-assisted communication systems. From the numerical results, we demonstrate that the proposed CNN-AE learning system shows significantly better BER performance for the RIS-assisted communication systems compared to the theoretical values.


\bibliographystyle{IEEEbib}

\bibliography{di}
\end{document}